%% file: main.tex
\documentclass{AUJarticle}
\input{config}
\input{macros}

\input{acronyms}

\begin{document}

\title{Dynamic System Model Generation for Online Fault Detection and Diagnosis of Robotic Systems}

\addauthor{Johannes Kohl, Georg Muck, Georg Jäger, Sebastian Zug}
{Freiberg University of Mining \& Technology, Institute for Computer Science,
	Akademiestra{\ss}e 6, 09599 Freiberg, Germany}
{\{johannes.kohl, georg.muck, georg.jaeger, sebastian.zug\}@informatik.tu-freiberg.de}

\issuev{35}
\issuen{1}
\issued{March 2014}

\shortauthor{Johannes Kohl, Georg Muck, Georg Jäger, Sebastian Zug}
\shorttitle{WIP: Monitoring for ROS-based Systems}

\thispagestyle{plain}

\maketitle

\begin{abstract}

	\input{00_abstract}

\end{abstract}

\ifdraft%
	\section{Conference Details}

	\textcolor{red}{%
		\begin{itemize}%
			\item Link: \url{http://www.ada-europe.org/conference2025/}%
			\item CfP: \url{http://www.ada-europe.org/conference2025/cfp.html#cfpwip}%
			\item Deadline: 24 February 2025%
			\item Page limit: 4%
		\end{itemize}%
	}
\fi


\makeatletter{\renewcommand*{\@makefnmark}{}\footnotetext{This work was partially funded by the German Federal Ministry for Digital and Transport within the mFUND program (grant no. 19FS2025A - Project Ready for Smart City Robots).}\makeatother}
%
\input{01_motivation}

\input{02_state_of_the_art}

\input{03_concept}

\input{04_evaluation}

\input{05_future_work}

\bibliographystyle{ieeetr}
\bibliography{bibliography}
\balance

\end{document}

%% file: config.tex
\usepackage[cmex10]{amsmath}
\usepackage[utf8x]{inputenc}
\usepackage[nocompress]{cite}
\usepackage{graphicx, multirow, booktabs, color, listings}
\usepackage{balance}
\usepackage{url}
\usepackage{hyperref}
\usepackage{cleveref}
\usepackage{todonotes}
\usepackage{ifthen}
\usepackage[inline]{enumitem}
\usepackage{amssymb}
\usepackage[printonlyused, nolist, withpage]{acronym}

\graphicspath{{figures/}}

\usepackage{tikz}
\usetikzlibrary{shapes.geometric,arrows.meta,calc,babel,backgrounds}
\tikzset{
    block/.style={
        rectangle,
        rounded corners,
        minimum width=2.5cm,
        inner sep=3pt,
        align=center,
        draw=black,
        fill=white
    },
    db/.style={
        cylinder,
        thick,
        draw = black,
        aspect = .085, 
        inner sep=5pt,
        minimum width = 1.5cm,
        shape border rotate = 90
    }
}

\newboolean{showcontent}
\setboolean{showcontent}{false} 

%% file: macros.tex
\newif\ifdraft
	\draftfalse

	\definecolor{ColorJK}{RGB}{100,209,255}
	\definecolor{ColorGM}{RGB}{190,174,255}
	\definecolor{ColorGJ}{RGB}{204, 153, 0}

	\newcommand{\itodo}[1]{
		\ifdraft
			\todo[inline]{\scriptsize\textbf{TODO:} #1}
		\fi}

%% file: acronyms.tex
\begin{acronym}
    \acro{mbse}[MBSE]{Model Based System Engineering}
    \acro{dshadow}[DS]{Digital Shadow}
    \acro{dtwin}[DT]{Digital Twin}
    \acro{dthread}[DThread]{Digital Thread}
    \acro{sysml}[SysML]{Systems Modelling Language}
    \acro{bit}[BIT]{Build-In Tests}
    \acro{sm}[SM]{System Modelling}
    
    \acro{stream}[STREAM]{Sensor and Telemetry: Robots Event-Based Aggregation and Monitoring}
    \acro{rf}[RF]{Robotics Framework}
    \acro{dsys}[DSys]{Distributed System}
    \acro{os}[OS]{Operating System}

    \acro{rc}[RC]{Root Cause}
    \acro{ft}[FT]{Fault Trajectory}
    \acro{rca}[RCA]{Root Cause Analysis}
    \acro{fdd}[FDD]{Fault Detection and Diagnosis}
\end{acronym}

%% file: 00_abstract.tex
With the rapid development of more complex robots, \acl{fdd} becomes increasingly harder.
Especially the need for predetermined models and historic data is problematic because they do not encompass the dynamic and fast-changing nature of such systems.
To this end, we propose a concept that actively generates a dynamic system model at runtime and utilizes it to locate root causes.
The goal is to be applicable to all kinds of robotic systems that share a similar software design.
Additionally, it should exhibit minimal overhead and enhance independence from expert attention.

%% file: 01_motivation.tex
\section{Motivation}
\label{sec:motivation}

Mobile robotic systems are increasingly deployed in open environments, requiring them to handle a variety of external influences.
This complexity increases the potential for diverse faults~\cite{khalastchi2018fault}, ranging from hardware and actuator failures to software and interaction-related issues.
Detecting these faults at design time is challenging due to the complexity of such systems.
However, undetected faults threaten the reliability and safety of robotic systems, making online \ac{fdd} essential~\cite{golombek2011fault}.

Conventional \ac{fdd}, however, struggles with the complexity of mobile robots and the dynamic nature of rapid prototyping.
This is due to the characteristics of continuous changes from this approach.
To address these challenges, this article describes a work-in-progress approach to dynamic \ac{fdd} for mobile robots, using a \acl{dshadow} -- an aggregated, passively generated data footprint that reflects the robotic system \cite{MBSE_DMDSDT}.
We aim at an \ac{fdd} design that minimizes computational overhead as well as expert knowledge and is applicable to a variety of robots.
For that reason, \Cref{sec:concept1} introduces a generalized mathematical meta-model to encompass a broad range of \acp{dsys} like ROS2.

In the following, we analyse \acfi{sm} for robotics in \Cref{subsec:system_modeling}.
Next, we review previous developments in \ac{fdd} and identify the necessity of an online system model in \Cref{subsec:fault_analysis}.
Based on these insights, we construct a mathematical meta-model that shadows a generic \ac{dsys} in \Cref{sec:concept1}, apply it to a software layer in \Cref{sec:concept2} and suggest how it can be utilized effectively in \Cref{sec:concept3}.
We conclude the article with an outlook on planned future work in \Cref{sec:future_work}.

%% file: 02_state_of_the_art.tex
\section{State of the Art on \ac{sm} and \ac{fdd}}
\label{sec:state_of_the_art}

To review the current state of \ac{fdd} and \ac{sm}, we focus on \ac{mbse} and \ac{rca}. This enables us to identify techniques aiming for minimal computational overhead and expert knowledge. Moreover, this allows the identification of requirements by assessing the shortcomings of the proposed methods.

Within this work, we use the following terminology:
A \textit{robot} is the physical platform hosting 
the \acfi{rf}, which describes a software infrastructure layer that handles communication, lifetime, etc. like ROS2 \cite{Macenski_2022_ros2} or ORCA \cite{makarenko2006orca}.
A \acfi{dsys} in turn describes the software stack built for the robot and is running in the \ac{rf}.
A \ac{dsys} consists of \textit{Components} which perform system tasks and generate process data, and \textit{Distributors} which facilitate the communication between components, e.g. Topics in ROS2 \cite{Macenski_2022_ros2}.
Furthermore, we follow the definitions of \cite{avizienis2001fundamental}:
A \textit{symptom} is an indicator of a failure, like service degradation or alarms.
A \textit{failure} is a deviation between actual and expected system behaviour.
An \textit{error} is the part of the system state that may cause a failure.
A \textit{fault} is a possible or real cause of an error.

\subsection{System Modelling}
\label{subsec:system_modeling}

Building on this foundation, we explore approaches of \ac{mbse} in robotics to develop a dynamic, data-driven model that reduces expert knowledge and enables the creation of a \acl{dshadow}.
As the goal is to systematically detect and diagnose faults
, \ac{mbse} provides a structured framework to formally model system behaviour, interactions and failure propagation \cite{MBSEValBenMod}.
It is an approach that enables the establishment of models for various use cases, ensuring their structure and maintainability even in complex scenarios\cite{MBSEValBenMod}\cite{mbseRamos11}.
This means MBSE supports the identification of faults at design time using a holistic view\cite{MBSEObtFaTrThSyDi}.
Existing work widely adopts \textit{Systems Modelling Language} as the modelling language for system representation due to its expressiveness and versatility \cite{mbseRamos11}.

Previous studies have largely focused on the interaction between hardware and software, e.g. \cite{MBSETaOpSyDrEn}, \cite{MBSEDevofAGVSystem} and \cite{MBSEEnSiFrRoGra}.
\cite{MBSEFDAFDI} highlights the role of \ac{bit}, which are embedded within the system architecture, predominantly during its design phase, to detect and diagnose failures in real time.
Thus, they maximize the number of detectable faults and enable more efficient maintenance by reducing diagnostic latency.
The author also notes the challenge of retrofitting \ac{bit} capabilities into already existing systems.
Since the targeted systems have already passed their design phase, the approach is not ideal.

An alternative is to use a \acl{dshadow}, which is a virtual representation that mirrors the physical system at runtime.
Contrary to a \acl{dtwin} \cite{MBSE_DMDSDT}, the automated data flow only reaches from the physical system to the virtual model and therefore is not bidirectional.
In conclusion, the \acl{dshadow} cannot manipulate the physical system as a \acl{dtwin} could. Both use simulations for the prediction of faults or maintenance needs (predictive maintenance).

\cite{MBSEIRDT} uses \ac{mbse} to construct a scalable \acl{dtwin} of industrial robots that is used to gather failure data and enable predictive maintenance.
Furthermore, the authors propose a modular, hierarchical and generic approach to enhance reusability, scalability, and interoperability across different industrial robot applications and lifecycle stages.
However, the approach describes the creation of a \acl{dtwin} throughout the entire lifecycle, whereas our approach targets systems beyond the design phase.

In summary, despite \ac{mbse}'s adoption in robotics, the complexity of modelling robotic systems leads to the lack of such a model \cite{MBSEDevofAGVSystem}.
To fill the gap, we propose the use of \ac{mbse} 

to create an abstract meta-model in \Cref{sec:concept1}.
It aims to enable the automated generation of system models at runtime and provide knowledge that can be used to detect and localize faults.

\subsection{Root Cause Analysis}
\label{subsec:fault_analysis}

\ac{fdd} is a generic term for inductive methods, postulating a fault and analysing its effect on the system, and deductive methods, postulating a failure and analysing which errors or faults may influence it \cite{Xing2008}.
\textit{Failure Mode and Effects Analysis} is one such inductive analysis method focusing on the system's risks.
It is usually conducted in the design phase and thus not well suited for automation.
\textit{Fault Tree Analysis} and \textit{Event Tree Analysis} are deductive and inductive respectively \cite{Xing2008} and analyse the system's causal connections.
They can be used at runtime to identify a \acfi{rc}, a fault that causes another but is not caused by one \cite{soleMRE17}.
Knowledge about \acp{rc} can be used to enable subsequent components to initiate appropriate countermeasures.

Such methods that aim to identify \acp{rc} from a set of symptoms are categorized as \ac{rca} \cite{soleMRE17}.
\ac{rca} generally relies on a \textit{System Object Taxonomy}, i.e. a graph modelling the targeted \ac{dsys} using objects, representing its components and distributors, as well as their interdependencies.
This is needed to retrieve a \acfi{ft}, i.e. an ordered set of system objects exhibiting symptoms, starting at the initially detected symptom and ending in a \ac{rc} candidate.
In order to detect these symptoms, many RCA algorithms rely on signals carrying information about the symptom, called \textit{Alerts}.
While no RCA algorithms for robotics were found in the literature review, related proposals exist in cloud computing and microservices \cite{OpenRCA} \cite{GRANO} \cite{MicroRCA}.
These algorithms are promising, as their approach of decomposing monolithic structures into modular services with lightweight communication mirrors the distributed architecture of \acp{rf}.

OpenRCA \cite{OpenRCA} identifies \acp{rc} by correlating symptoms in time series of alerts, leveraging multiple methods to balance their strengths and weaknesses. It maps symptoms to a system topology for tracing \acp{ft}.

\cite{GRANO} proposes GRANO (\textbf{Gra}ph \textbf{Ano}maly) which aims to score each system topology component with a \textit{Root Cause Relevance}, i.e. a weight to get taken into account when reviewing \acp{ft}.
It is calculated based on the amount and severity of alerts emitted for each component.
While this allows an efficient search for \acp{rc} it still performed manually.

Another approach is taken by MicroRCA \cite{MicroRCA} which only looks at communication timing and service performance data.
These metrics are analysed for symptoms like anomalous CPU utilization to be annotated into a system topology graph.
This annotated graph in turn is analysed by the Personalized PageRank \cite{personalizedPageRank} algorithm, sorting inputs based on their importance in order to find \acp{rc}.

Even though the above approaches show promising evaluation results, their strategies have two main deficiencies in regard to the stated requirements of this article.
Firstly, they expect the \ac{rf} to supply facilities that emit alerts on detected symptoms.
Since this cannot be done without expert knowledge, we propose the use of a plugin system to detect faults.
Secondly, OpenRCA and GRANO require a holistic system object taxonomy, assumed to be statically provided by expert knowledge.

To align with our autonomy goals and to reduce expert knowledge, a meta-model to dynamically build and maintain a graph of the distributed system is defined in the next section.

%% file: 03_concept.tex
\section{Concept}
\label{sec:concept}

For clarity, our approach is thematically subdivided into the generation of the system model (\Cref{sec:concept2}) and failure analysis (\Cref{sec:concept3}). The former focuses on modelling the system to support FDD. The latter aims to utilize the created model to extract the \ac{ft}.
The underlying conceptual meta-model is described in \Cref{sec:concept1}.

\subsection{System Model}\label{sec:concept1}

To ensure a structured yet adaptable representation of a \ac{dsys}, we propose a meta-model that abstracts system components.

In this work, we model a \ac{dsys} as a directed graph $G=(M,E,\theta)$. The graph consists of a set of members $M=M_A\cup M_P$, where $\:M_A \cap M_P=\emptyset$, and a set of not disjoint edges $E = E_0 \cup E_i$ with $0<i<N\in\mathbb{N}_0$. The attribute function is defined as $\theta=\theta_A\cup\theta_P$, with $\theta_A\colon M_A\to A_A\subseteq \mathbb{R}^a$ and $\theta_P\colon M_P\to A_P\subseteq\mathbb{R}^p$, where $a,p\in \mathbb{N}$.

The system differentiates between two types of members.
Active members execute or control operations, e.g. Nodes in ROS2 \cite{Macenski_2022_ros2}, while passive members enable data distribution or serve as organisational units, e.g. Topics in ROS2 \cite{Macenski_2022_ros2}.

The graph contains multiple types of communication edges, defined as $E_0=E_{send}\cup E_{pub}\cup E_{sub}$.
Specifically, sending edges are given by $E_{send} = (M_A\times M_A) \times \{0\}$, publishing edges by $E_{pub} = (M_A\times M_P) \times \{0\}$, and subscribing edges by $E_{sub} = (M_P\times M_A) \times \{0\}$.
Each edge represents a one-to-one connection, but since passive members act as intermediaries, they enable many-to-many connections between active members $M_A$.
The edge-induced subgraph $G_0=(M_0,E_0,\theta)$ represents the communication structure of the distributed system,
where $\forall m, n\in M_0 \subseteq M: (m, n) \in E_0 \wedge (n,m) \in E_0$.
Additional edges model other relationships and are defined as $E_i = (M\times M)\times {i}$.
The corresponding edge-induced subgraph $T_i=(M_i,E_i,\theta)$ of $G$ forms a rooted tree, where the edges are given by $E_i = \{(v,u)\vert ((v,u),i) \in E\}$, ensuring that $T_i$ contains only the vertices $M_i$, with $M_i$ formed analogously to $M_0$ above.

\subsection{Data Aggregation}\label{sec:concept2}
Building on the established meta-model, we now focus on modelling conceptual data aggregation.
Using a top-down approach, we enable hierarchical observation, allowing broad system analysis that becomes more precise as needed.

\subsubsection{Top-Down Aggregation for \ac{fdd}}\label{subsubsec:tree}
The top-down approach is characterized by its initiation at the abstract levels, progressing in an incremental manner towards more detailed levels. The approach can be applied by using the edge-induced trees described in \Cref{sec:concept1}. Let $C_i(v)=\{u\in M_i|(v,u) \in E_i\}$ denote the children of $v \in M_i$. The accumulated property function $\psi_i(v)\colon M_i \to A$ is then recursively defined as $\psi_i(v)=\sum_{c\in C_i(v)}\left(\psi_i(c) + \theta(c)\right)$. This recursive formulation closely resembles a depth-first search.

Building on the meta-model (\Cref{sec:concept1}) and the top-down approach described above, we now focus on projecting the conceptual model onto the software layer.

\subsubsection{Projection to the Software Layer} \label{subsubsec:metamodeltosoftwarelayer}
In this approach, the distributed system described in \Cref{sec:state_of_the_art} runs on an \ac{os} and consists of \textit{components} and \textit{distributors}. The \ac{os} is expected to provide the necessary infrastructure, resources, and services for execution. In this work, we focus on the Unix family of \ac{os}s.

A component corresponds to an active Member $m_{0,A}\in M_{0,A}= M_0\cap M_A$ in the edge-induced subgraph $G_0$ on a software layer. It has a distinct name and can be associated with a process on the \ac{os}. Passive Members $m_{0,P}\in M_{0,P}=M_0\cap M_P$ in $G_0$ serve as message distributors and are termed distributors. Communication occurs via the edges $E_0$.

After projecting the meta-model onto the software layer, the next sections will examine various aggregation techniques.

\subsubsection{Fundamental Aggregation}\label{subsubsec:FundAgg}

The fundamental approach involves establishing relationships between components and distributors, 
mapping the directed connections between individual components including associated distributors. These connections enable querying the Graph for all components $r\in G_0$ predecessors $M_0\ni r_{pre}=\{m_0\in M_0|\exists E_0\ni e_0=(m_0,r)\}$, active predecessors $M_{0,A}\ni r_{pre}=\{m_{0,A}\in M_{0,A}|\exists E_0\ni e_0=(m_{0,A},r)\}$ and passive predecessors $M_{0,P}\ni r_{pre}=\{m_{0,P}\in M_{0,P}|\exists E_0\ni e_0=(m_{0,P},r)\}$
and successors in an analogue way.
This enables subsequent FDDs to derive a submodel containing the necessary information.

Another perspective will be introduced next by using a tree $T_i$.

\subsubsection{Process Tree as a top-down approach}\label{subsubsec:processTree}
In a Unix-based \ac{os}, all processes except the root process are linked to a parent, forming a process tree. Since related components and their processes are typically grouped during their startup (e.g., via initialization scripts), they share common ancestors. This tree structure can be created by tracing the parent processes of $M_A$. If multiple root processes exist, a common virtual root can be introduced.

The provided structures described in \Cref{subsubsec:FundAgg} and \Cref{subsubsec:processTree} are used to detect faults in the following section.

\subsection{Fault Detection and Root Cause Analysis}
\label{sec:concept3}
In this section, we present a concept to address the problems identified in \Cref{subsec:fault_analysis}. It leverages a dynamic subgraph of the system model and proposes a symptom detection method to generate alerts, ensuring independence from the \acp{rf}'s provision of such a facility.

In the following, we define the subgraph $G_j = (M_j, E_j)$ with members $M_j \subseteq M$, edges $E_j \subseteq E: \forall ((m_0,m_1), 0) \in E_j\colon m_0,m_1 \in M_j$ and iteration $j \in \mathbb{N}_0$. Here, an iteration describes a stage of the subgraph's construction.
\itodo{Add: "As the concept works on the software layer, active members may be called components and passive members may be called distributors." ?}

Next, the in \Cref{subsubsec:symptomDetection} explained symptom detection is part of the dynamic allocation of the subgraph in \Cref{subsubsec:dynamicSubgraph}, which in turn is used to extract \acp{ft} in \Cref{subsubsec:faultTrajectory}.

\subsubsection{Symptom Detection}
\label{subsubsec:symptomDetection}

As discussed in \Cref{subsec:fault_analysis}, providing a comprehensive symptom detection is beyond the scope of this article, making reliance on expert knowledge unavoidable. To use this knowledge efficiently, we propose a plugin system to be populated by the expert, which will emit alerts for detected symptoms.

For the purpose of providing information to the symptom detection, we define the watchlist $W_j \subseteq M$ which provides the attributes $\theta(w \in W_j)$ to the plugins.
Alerts emitted by the symptom detection get stored in the alert database saving their origin and a timestamp, in order to extract and analyze time series of alerts per member later.

\subsubsection{Dynamic Subgraph Allocation}
\label{subsubsec:dynamicSubgraph}

The alerts emitted by the symptom detection also are used to expand the subgraph.
This means that when an alert gets emitted for member $m_{alert} \in W_j$, the sets for the next iteration will be defined as follows:
\begin{enumerate*}[label=\Alph*)]
	\item $M_{i+1} = M_j \cup \{m_{alert}\}$,
	\item $E_{i+1} = E_j \cup \{((m, m_{alert}), 0) \in E| m \in M_j\}$ and
	\item $W_{i+1} = W_j \cup \{m \in M | ((m_{alert}, m), 0) \in E\}$.
\end{enumerate*}

Iteration $i = 0$ reflects a special case: since the watchlist is empty, nothing can be added to the subgraph.
Thus, an initialization is needed, which can be done in two ways.
Firstly, by directly populating the watchlist with members from a configuration based on factors like importance or failure probability.
Secondly, by identifying components in the process tree, extracted from $G$ in \Cref{subsubsec:processTree}, with anomalous metrics using process monitoring.

\subsubsection{\acl{ft} Extraction}
\label{subsubsec:faultTrajectory}

At iteration $j_{extract}$, the subgraph $G_{j_{extract}}$ can be used to trace a \ac{ft} and by extension a \ac{rc}.
An assortment of different symptom correlation methods can be used in conjunction to trace \acp{ft}.
One such method is Symptom Co-Occurrence which aims to detect if two sets of time series of alerts co-occurred, using an Iterative Closest Points algorithm adapted to one-dimensional space.
Another method is Symptom Time Lag Analysis that aims to detect symptom pairs by comparing their time lag to a time lag distribution.
The retrieved \acp{ft} are being ranked by the average strength of their dependencies and length \cite{OpenRCA}.

%% file: 04_evaluation.tex

%% file: 05_future_work.tex
\section{Conclusion and Future Work}
\label{sec:future_work}

Within this work we defined a mathematical meta-model that aims to universally represent \aclp{dsys} in order to facilitate dynamic \acl{fdd} in the context of rapid prototyping.

We endeavour to expand our present model.
This enhancement will facilitate a more comprehensive analysis of network activity.
Additionally, we will investigate methods to ensure secure and efficient data aggregation and synchronization from various clients, addressing potential challenges related to data synchronization.

Furthermore, the RCA algorithm needs to be finalized and refined.
This especially necessitates more research about alternative symptom correlation mechanisms and possible automations of the initialization.
This also implies testing the feasibility of correlation methods established in \Cref{subsec:fault_analysis} for online usage or the addition of an offline component to aid the live execution.
Another goal would be adding functionality to differentiate between permanent and non-permanent faults and handle them separately.

The subsequent steps will be to focus on the aforementioned points and to implement the pipeline in ROS2.
Based on the implementation the pipeline will be tested in multiple real-world scenarios and simulations to evaluate its applicability on a broad range of robotic systems.

%% file: main.bbl
\begin{thebibliography}{10}

\bibitem{khalastchi2018fault}
E.~Khalastchi and M.~Kalech, ``On fault detection and diagnosis in robotic
  systems,'' {\em ACM Comput. Surv.}, vol.~51, Jan. 2018.

\bibitem{golombek2011fault}
R.~Golombek, S.~Wrede, M.~Hanheide, and M.~Heckmann, ``Online data-driven fault
  detection for robotic systems,'' in {\em 2011 IEEE/RSJ International
  Conference on Intelligent Robots and Systems}, pp.~3011--3016, 2011.

\bibitem{MBSE_DMDSDT}
V.~Lopez and A.~Akundi, ``A conceptual model-based systems engineering (mbse)
  approach to develop digital twins,'' in {\em 2022 IEEE International Systems
  Conference (SysCon)}, pp.~1--5, 2022.

\bibitem{Macenski_2022_ros2}
S.~Macenski, T.~Foote, B.~Gerkey, C.~Lalancette, and W.~Woodall, ``Robot
  operating system 2: Design, architecture, and uses in the wild,'' {\em
  Science Robotics}, vol.~7, May 2022.

\bibitem{makarenko2006orca}
A.~Makarenko, A.~Brooks, and T.~Kaupp, ``Orca: Components for robotics,'' in
  {\em International Conference on Intelligent Robots and Systems (IROS)},
  pp.~163--168, Citeseer, 2006.

\bibitem{avizienis2001fundamental}
A.~Avizienis, J.-C. Laprie, B.~Randell, {\em et~al.}, ``Fundamental concepts of
  dependability,'' {\em Technical Report Series-University of Newcastle upon
  Tyne Computing Science}, 2001.

\bibitem{MBSEValBenMod}
K.~Henderson and A.~Salado, ``Value and benefits of model-based systems
  engineering (mbse): Evidence from the literature,'' {\em Systems
  Engineering}, vol.~24, no.~1, pp.~51--66, 2021.

\bibitem{mbseRamos11}
A.~L. Ramos, J.~V. Ferreira, and J.~Barcel{\'o}, ``Model-based systems
  engineering: An emerging approach for modern systems,'' {\em IEEE
  Transactions on Systems, Man, and Cybernetics, Part C (Applications and
  Reviews)}, vol.~42, no.~1, pp.~101--111, 2011.

\bibitem{MBSEObtFaTrThSyDi}
A.~H. de~Andrade~Melani and G.~F.~M. de~Souza, ``Obtaining fault trees through
  sysml diagrams: A mbse approach for reliability analysis,'' in {\em 2020
  Annual Reliability and Maintainability Symposium (RAMS)}, pp.~1--5, IEEE,
  2020.

\bibitem{MBSETaOpSyDrEn}
L.~Brisacier-Porchon and O.~Hammami, ``Tackling optimization and system-driven
  engineering in coupling physical constraints with mbse: The case of a mobile
  autonomous line of products,'' in {\em Conference on Systems Engineering
  Research}, pp.~441--459, Springer, 2023.

\bibitem{MBSEDevofAGVSystem}
K.~Aloui, M.~Hammadi, A.~Guizani, T.~Soriano, and M.~Haddar, ``Development of
  an agv system using mbse method and multi-agents’ technology,'' in {\em
  International Conference Design and Modeling of Mechanical Systems},
  pp.~103--114, Springer, 2021.

\bibitem{MBSEEnSiFrRoGra}
P.~K.~M. Sekar and J.~S. Baras, ``Model-based systems engineering simulation
  framework for robot grasping,'' in {\em INCOSE International Symposium},
  vol.~32, pp.~82--89, Wiley Online Library, 2022.

\bibitem{MBSEFDAFDI}
N.~Mahmood, S.~Cimtalay, and D.~N. Mavris, ``Model based systems engineering
  (mbse) applied to fault detection analysis of vehicle subsystems,'' in {\em
  2022 IEEE Aerospace Conference (AERO)}, pp.~1--11, IEEE, 2022.

\bibitem{MBSEIRDT}
X.~Zhang, B.~Wu, X.~Zhang, J.~Duan, C.~Wan, and Y.~Hu, ``An effective mbse
  approach for constructing industrial robot digital twin system,'' {\em
  Robotics and Computer-Integrated Manufacturing}, vol.~80, p.~102455, 2023.

\bibitem{Xing2008}
L.~Xing and S.~V. Amari, {\em Fault Tree Analysis}, pp.~595--620.
\newblock London: Springer London, 2008.

\bibitem{soleMRE17}
M.~Sol{\'{e}}, V.~Munt{\'{e}}s{-}Mulero, A.~I. Rana, and G.~Estrada, ``Survey
  on models and techniques for root-cause analysis,'' {\em CoRR},
  vol.~abs/1701.08546, 2017.

\bibitem{OpenRCA}
B.~Żurkowski and K.~Zieliński, ``Root cause analysis for cloud-native
  applications,'' {\em IEEE Transactions on Cloud Computing}, vol.~12, no.~1,
  pp.~232--250, 2024.

\bibitem{GRANO}
H.~Wang, P.~Nguyen, J.~Li, S.~Kopru, G.~Zhang, S.~Katariya, and
  S.~Ben-Romdhane, ``Grano: interactive graph-based root cause analysis for
  cloud-native distributed data platform,'' {\em Proc. VLDB Endow.}, vol.~12,
  pp.~1942--1945, Aug. 2019.

\bibitem{MicroRCA}
L.~Wu, J.~Tordsson, E.~Elmroth, and O.~Kao, ``Microrca: Root cause localization
  of performance issues in microservices,'' in {\em NOMS 2020 - 2020 IEEE/IFIP
  Network Operations and Management Symposium}, pp.~1--9, 2020.

\bibitem{personalizedPageRank}
G.~Jeh and J.~Widom, ``Scaling personalized web search,'' in {\em Proceedings
  of the 12th International Conference on World Wide Web}, WWW '03, (New York,
  NY, USA), p.~271–279, Association for Computing Machinery, 2003.

\end{thebibliography}
